# PDC – a probabilistic distributional clustering algorithm: a case study on suicide articles in PubMed

Rezarta Islamaj, PhD, Lana Yeganova, PhD, Won Kim, PhD, Natalie Xie,
W. John Wilbur, MD, PhD, Zhiyong Lu, PhD

National Library of Medicine, National Institutes of Health, Bethesda MD, USA

The need to organize a large collection in a manner that facilitates human comprehension is crucial given the ever-increasing volumes of information. In this work, we present PDC (probabilistic distributional clustering), a novel algorithm that, given a document collection, computes disjoint term sets representing topics in the collection. The algorithm relies on probabilities of word co-occurrences to partition the set of terms appearing in the collection of documents into disjoint groups of related terms. In this work, we also present an environment to visualize the computed topics in the term space and retrieve the most related PubMed articles for each group of terms. We illustrate the algorithm by applying it to PubMed documents on the topic of suicide. Suicide is a major public health problem identified as the tenth leading cause of death in the US. In this application, our goal is to provide a global view of the mental health literature pertaining to the subject of suicide, and through this, to help create a rich environment of multifaceted data to guide health care researchers in their endeavor to better understand the breadth, depth and scope of the problem. We demonstrate the usefulness of the proposed algorithm by providing a web portal that allows mental health researchers to peruse the suicide-related literature in PubMed.

## 1. Introduction

The rapid growth of the biomedical literature in PubMed can make it challenging for researchers, clinicians, healthcare providers and the general public to find the information they need: an average search in PubMed returns hundreds to thousands of documents. Studies have shown that physicians, for example, need to find relevant information at the point of care, driving the need for human comprehension of any large document collection. Providing access to a literature collection in a way that is intuitive, organized and easy to comprehend is crucial for clinical decision making.

Advances in technology enable large-scale data analysis to process and extract useful information from published literature or social media that pertain to a given topic or a subject term. Many studies have tried to capture and organize in a humanly understandable format the body of scientific literature in PubMed, and other big collections to provide a global view of a field. Research methods have been developed to organize the PubMed literature into meaningful clusters to address specific questions. For example, (Guo and Laidlaw 2015) combine established coherent topics discovered by topic models and concept map analysis to encourage exploration and research idea generation. MeSH terms have traditionally been used (Srinivasan 2001) and continue to be used to visualize and present research topics in PubMed (Kim, Yeganova et al. 2016, Yang and Lee 2018). GoPubMed (Doms and Schroeder 2005) utilizes Gene Ontology (GO) terms to categorize and group PubMed publications. Semantic Medline (Rindflesch, Kilicoglu et al. 2011) provides literature search and knowledge exploration by the summarization of the semantics of biomedical documents. More recently, (Ilgisonis, Lisitsa et al. 2018) proposed concept-centered semantic maps of PubMed publications, by creating concept links based on semantic similarity between two concepts. (Yeganova, Kim et al. 2018) proposed an algorithm for discovering themes in biomedical literature and apply it to analyze a collection of articles on the topic of single nucleotide polymorphisms. (Wu, Jin et al. 2017), used co-occurrence analysis to investigate trends in psychiatry and (Wang, Ding et al. 2016) researched topics in literature on adolescent substance abuse and depression.

Numerous studies on topic analysis can be found in the computer science literature. For example, (Blei, Ng et al. 2003) have developed the well-known Latent Dirichlet Allocation (LDA), an unsupervised learning method to extract topics from a corpus, which models topics as a multinomial distribution over words. Since its introduction, LDA has been extended and adapted to several applications. For example, the Correlated Topic Model, introduced by (Blei and Lafferty 2007), uses the logistic normal distribution instead of the Dirichlet to address the issue of modelling correlations between topics, which LDA does not. The hierarchical LDA, developed by (Griffiths, Jordan et al. 2004), groups topics together in a hierarchy. (He, Chen et al. 2009) combined LDA and citation networks to address the problem of topic evolution. However, one common problem with LDA, as with many other methods, is the need to decide a priori the value for the parameter indicating the desired number of topics. When dealing with large amounts of life-scale data, users generally have no anticipated number of clusters in mind. Results, however, can change substantially depending on the parameter chosen.

In this work, we describe PDC, a novel clustering algorithm and apply it to explore publications in PubMed. The proposed algorithm is generic and may be applied to any collection of documents. The algorithm uses a mathematically defined optimization criterion that naturally produces a set of topics. Using the PDC, we identify a partitioning of a term set into disjoint groups of closely related terms (single terms, pairs of terms, and MeSH terms) that define the topics within a selected subject query in PubMed. We further provide a visualization environment that not only allows one to observe the global landscape of the selected subject query but also to explore each cluster by providing access to topic terms and PubMed articles most related to the identified clusters.

We utilize the PDC algorithm in the scope of suicide related literature in PubMed. A significant amount of work has been attributed to computational approaches developed to address problems of mental health and suicide. For example, (Yates, Cohan et al. 2017) propose methods for identifying posts in support communities that may indicate a risk of self-harm. Authors in (De Choudhury, Kiciman et al. 2017) develop a statistical methodology to infer which individuals could undergo transitions from mental health discourse to suicidal ideation. With the application of the PDC on the suicide related literature, and our computational visualization of the literature pertaining to suicide, we do not claim to offer solutions, rather we aim to present the published data as it partitions naturally following this probabilistic distributional approach and open up the results to clinicians and researchers to help them visualize potential areas of interest.

This study contributes on two dimensions. First, the PDC algorithm represents an algorithmic contribution of a novel method for finding topics from large amounts of literature. Second, we provide an extensive analysis of suicide literature in PubMed: to our knowledge, this is the first study attempting to analyze the suicide literature in PubMed.

The rest of the paper is organized as follows. In Section 2, we describe in detail our clustering approach and the framework we develop for visualizing these results. In Section 3 we apply PDC to analyze the ~81,000 PubMed documents retrieved with query "suicide", demonstrate the computed topics and topic terms, and show how we propose browsing PubMed articles retrieved with topic terms. In Section 4, we discuss our clustering approach and draw conclusions.

## 2. Methods

### 2.1. *A Probabilistic Clustering Formulation*

A general clustering problem can be defined as follows: let us suppose we are given a nonempty finite set of objects, $U$, and a probability function $p$ with the interpretation that for any objects $x, y \in U$, $p(x, y)$ has the interpretation as the probability that $x$ and $y$ should be clustered together. As such, we require $p(x, y)$ to be a symmetric function. Consider the functions

$$\delta : U \times U \to \{0, 1\} \quad (1)$$

that satisfy the pseudo metric axioms

$$\begin{aligned} &\delta(x, x) = 0, \ \forall x \in U \\ &\delta(x, y) = \delta(y, x), \ \forall x, y \in U \\ &\delta(x, z) \leq \delta(x, y) + \delta(y, z), \ \forall x, y, z \in U. \end{aligned} \quad (2)$$

We will refer to functions satisfying (1) and (2) as partition functions as it is simple to show that they are in one-to-one correspondence with partitions of the set $U$ into disjoint subsets or equivalently hard clusterings of the set $U$. Given any such partition function $\delta$, we can define its probability by

$$p(\delta) = \prod_{x, y \in U} p(x, y)^{(1 - \delta(x, y))} (1 - p(x, y))^{\delta(x, y)}. \quad (3)$$

Optimal clustering corresponds to finding a partition that maximizes the probability:

$$\delta' = \underset{\text{partition } \delta}{\arg\max}\ p(\delta) \qquad (4)$$

By applying log to (4) and dropping a term that does not involve $\delta$, we may rewrite (4) as

$$\delta' = \underset{\text{partition } \delta}{\arg\max}\ \sum_{x,y \in U \ni \delta(x,y)=0} \log\left(\frac{p(x,y)}{1-p(x,y)}\right). \qquad (5)$$

This is a typical formulation for the set partitioning problem and is known to be NP hard. Many heuristic approaches exist, depending on details. Our approach takes advantage that the sum in (5) only involves pairs of points from the same cluster, this will greatly speed up the search for the optimal clustering.

The PDC algorithm starts with all points in one cluster and involves calling a *splitting* algorithm repeatedly on each produced cluster to split all the clusters produced as far as possible. When no cluster can be further split to increase the sum in (5), we will have achieved a local optimum. This local optimum is the output of the algorithm.

The splitting algorithm examines each element to see how negatively it is related to other elements (negative log odds in (5)) and chooses the $k$ most negative elements for further analysis. The negativity of an element is measured as the sum of all the negative log odds it has with other elements in its cluster. The more polarized the relationships, the more advantage there may be in splitting a cluster. There can be no advantage in splitting unless there are negative relationships between elements. In all the work reported here we have used the value 10 for $k$, and we try all 10 options and keep the best result produced. The splitting algorithm heavily relies on *single point optimization*, a building block of our approach. Starting with a given split, single point optimization attempts to improve it by moving each element to a different cluster that most improves the sum. When all the elements have been tested, if at least one move improved the sum, the algorithm tries another pass through the data to check if the sum can be improved again. This continues if the sum improves but is limited to at most $m$ passes through the data. Big improvements in the sum generally come early in the computation and the limit is to avoid long calculations with almost no benefit. For the applications reported here we use $m=30$. The detailed c style pseudocode for the PDC algorithm including the splitting algorithm and single point optimization are provided in the Appendix.

The PDC algorithm is generic and can be applied to any collection of objects. Our interest is in applying it to a collection of documents on the topic of suicide that we seek to analyze. The results of the algorithm heavily rely on the definition of the probability function, which, in our application, represents the probability of two terms being related. In the next section we define how we compute that probability for pairs of terms $s$ and $t$.

## 2.2. Distributional Clustering probabilities

Let's assume that $s$ and $t$ represent two terms that occur in a set $V$ of documents of size $N$, and $n_s$ and $n_t$ represent the number of documents in $V$ that contain $s$ and $t$, respectively. We then consider whether $s$ and $t$ are *related*, i.e., whether they co-occur in documents in $V$ at a level higher than would be expected by chance. If we allow $r$ to denote such a relationship we seek an estimate for $p(r|data)$ where $data$ denotes the frequencies $N$, $n_s$, $n_t$, and $n_{st}$ and the latter is the number of documents containing both $s$ and $t$. Our estimate will be for

$$\log\left(\frac{p(r|data)}{1-p(r|data)}\right) = \log\left(\frac{p(r|data)}{p(\neg r|data)}\right) = \log\left(\frac{p(data|r)p(r)}{p(data|\neg r)p(\neg r)}\right) \qquad (6)$$

We applied Bayes theorem to obtain the term on the right. We will begin with the assumption that

$$p(\neg r) = p(r) = 0.5 \qquad (7)$$

so that we can ignore priors. We note that the set $V$ is naturally partitioned by $s$ and $t$ into four subsets $V_{s \wedge t}$, $V_{s \wedge \neg t}$, $V_{\neg s \wedge t}$, and $V_{\neg s \wedge \neg t}$ and these subsets naturally define four probabilities $p_{s \wedge t}$, $p_{s \wedge \neg t}$, $p_{\neg s \wedge t}$, and $p_{\neg s \wedge \neg t}$. Using these probabilities, we can write the probability of seeing the numbers $N$, $n_s$, $n_t$, and $n_{st}$, as:

$$MC p_{s \wedge t}^{n_{st}} p_{s \wedge \neg t}^{(n_s - n_{st})} p_{\neg s \wedge t}^{(n_t - n_{st})} p_{\neg s \wedge \neg t}^{(N - n_s - n_t + n_{st})} \qquad (8)$$

where $MC$ represents the appropriate multinomial coefficient. We will refer to this as the multinomial model and to (8) as the multinomial estimate. Since the four basic probabilities must sum to one the model involves the estimation of three unknowns. A slightly simpler model is based on the four sets $V_s$, $V_{\neg s}$, $V_t$, and $V_{\neg t}$. These sets also give rise to corresponding probabilities, but we only need estimate $p_s$ and $p_t$ because of the relations between them. We will refer to this as the binary independence model. Based on the binary independence model we can also estimate the probability of seeing the numbers $N$, $n_s$, $n_t$, and $n_{st}$ as

$$MC[p_s p_t]^{n_{st}} [p_s(1-p_t)]^{(n_s - n_{st})} [(1-p_s)p_t]^{(n_t - n_{st})} [(1-p_s)(1-p_t)]^{(N - n_s - n_t + n_{st})} \quad (9)$$

We note that the binary independence model approximates the multinomial model and in fact gives the correct estimates for the probabilities of the four sets $V_{s \wedge t}$, $V_{s \wedge \neg t}$, $V_{\neg s \wedge t}$, and $V_{\neg s \wedge \neg t}$ if and only if the independence condition is satisfied:
$$p_{s \wedge t} = p_s p_t. \quad (10)$$

We seek estimates for $p(data | r)$ and $p(data | \neg r)$. Our problem naturally breaks into two cases. The first case:

$$n_{st} \geq n_s n_t / N. \quad (11)$$

In order to estimate $p(data | r)$ we assume there may be a bias in that $s$ and $t$ may occur together more often than expected by chance. Since this dependence can only be captured by the multinomial model we take (8) to represent $p(data | r)$. In order to estimate $p(data | \neg r)$ we assume that any apparent bias in the co-occurrence of $s$ and $t$ is simply the result of a random process and that their distribution is appropriately modeled by the binary independence model. This leads us to take (9) to represent $p(data | \neg r)$. The interpretation here is that $s$ and $t$ are independent of each other and even if their overlap is large that must be understood as a random event. In the case of (11) then we have

$$\log\left(\frac{p(data | r)}{p(data | \neg r)}\right) = n_{st} \log\left(\frac{p_{s \wedge t}}{p_s p_t}\right) + (n_s - n_{st}) \log\left(\frac{p_{s \wedge \neg t}}{p_s(1 - p_t)}\right) + \\ + (n_t - n_{st}) \log\left(\frac{p_{\neg s \wedge t}}{(1 - p_s) p_t}\right) + (N - n_s - n_t + n_{st}) \log\left(\frac{p_{\neg s \wedge \neg t}}{(1 - p_s)(1 - p_t)}\right) \quad (12)$$

In the second case we have

$$n_{st} < n_s n_t / N \quad (13)$$

and we must reason differently. We again need an estimate for

$$\log\left(p(r | data) / (1 - p(r | data))\right) \quad (14)$$

and we make use of (6). Now, however, we interpret $p(data | r)$ with the assumption that $s$ and $t$ are related and we should have seen data consistent with (11) and the data we see, i.e. (13), is just a random accident. For the probability of this accident we use (9) as the appropriate estimate. In order to compute $p(data | \neg r)$ we assume there is a bias, only now against $s$ and $t$ occurring together. Since this negative dependence cannot be modeled by the binary independence model we must model it with the multinomial model and we estimate $p(data | \neg r)$ with (8). Notice how roles have been reversed. In this case (12) is replaced by the same equation with the sole difference being the right side of (12) has its sign switched. If we let $\Sigma_m$ represent the multinomial distribution and $\Sigma_i$ the binary independence distribution over the event space $\{s \wedge t, s \wedge \neg t, \neg s \wedge t, \neg s \wedge \neg t\}$ as represented in (8) and (9), then we can express our results using the Kullback-Liebler (KL) divergence (Kullback and Leibler 1951)

$$\log\left(\frac{p(data | r)}{p(data | \neg r)}\right) = \begin{cases} N \cdot D_{KL}(\Sigma_m \| \Sigma_i), & \text{if } n_{st} \geq n_s n_t / N \\ -N \cdot D_{KL}(\Sigma_m \| \Sigma_i), & \text{if } n_{st} < n_s n_t / N \end{cases} \quad (15)$$

This is helpful because we know the KL divergence is always positive unless the two distributions are identical, i.e., unless we have independence as defined by (10), in which case the KL divergence is zero.

Finally, we note that regardless of the case (6) can be written as

$$\log\left(\frac{p(r\mid data)}{1-p(r\mid data)}\right) = \log\left(\frac{p(data\mid r)}{p(data\mid \neg r)}\right) + \log\left(\frac{p(r)}{p(\neg r)}\right). \quad (16)$$

Thus, we can always adjust all our odds ratios up or down by a constant factor

$$\log\left(p(r)/p(\neg r)\right) \quad (17)$$

reflecting a prior log odds ratio different than the 0 produced by equation (7). In many practical applications it is important to set the prior log odds, (17), equal to some negative constant. This is because in the forgoing development we have set no lower bound on how related $s$ and $t$ must be to be considered related. Thus (11) may be an inequality by the tiniest of margins and still the left side of (16) will be positive and the algorithm will attempt to cluster $s$ and $t$ together. Note that the left side of (16) will be positive exactly when

$$\log\left(\frac{p(data\mid r)}{p(data\mid \neg r)}\right) > -\log\left(\frac{p(r)}{p(\neg r)}\right) \quad (18)$$

Thus setting (17) to a negative constant effectively sets a lower bound to the relatedness of terms the clustering algorithm will attempt to cluster together.

### 2.3. *Graphical Literature Analysis*

We have applied the distributional cluster analysis to terminology that occurs in subsets of MEDLINE records in the PubMed database. We illustrate the algorithm on a set of documents retrieved from PubMed with a query: *suicide OR suicide [MeSH Terms]*. The query retrieves ~81,000 documents that we denote as the set $V$ and let $M$ represent the whole PubMed. Take $W$ to be the set of single terms, term bigrams, and MeSH terms that appear in the titles and abstracts of documents in $M$. We can analyze each of these terms to see how it is distributed in records in $V$ and in $M-V$. If a term appears more in $V$ than expected by chance given its number of occurrences throughout $M$, we can compute a *p*-value that the term would appear this many or more times in elements of $V$ using the hypergeometric distribution. We then apply the Benjamini-Hochberg procedure (Benjamini and Hochberg 1995) with a false discovery rate set to 0.01. Since this may yield too many terms for practical analysis, we also apply a frequency limit as needed to trim down the size of the resulting set of terms. We denote the set of terms by $U$ and apply to it the PDC.

We begin our analysis of a set $U$ of terms by running the PDC algorithm with (17) set to zero to produce a clustering which we denote by $C_0$. Then if $C_n$ represents the clustering produced when (17) equals $-n/2$, we produce $C_{n+1}$ from $C_n$ by applying PDC with (17) set to $-(n+1)/2$ to all the clusters in $C_n$ of size greater than 100. Because more splitting takes place as (17) becomes more negative we progressively reduce the size of the large clusters until all clusters are of size 100 or less. The result is a set of $K$ clusterings $\{C_n\}_{n=0}^{K-1}$ at $K$ different levels (0 up to $K-1$) which are progressively finer the higher the level. The next step is to run through the clusters at each level, beginning with the lowest level and collect the clusters at each level that do not occur at any lower level. In this process we ignore any cluster of size one or that has an associated score of 0. We denote the resultant collection of clusters by $\bar{C}$. In the process of collecting the clusters we also number them beginning at 1 so we can write $\bar{C}=\{x_i\}$. Any cluster in $\bar{C}$ at a level above 0 will occur as a subset of some larger cluster in $\bar{C}$ at a lower level. This allows us to define a function $L:\bar{C}\to\bar{C}$ by setting $L(x)=x$ if $x\in C_0$ and $L(x)=y$ if $x\in\bar{C}\cap C_k$, $y\in\bar{C}\cap C_{k'}$, $k'<k$ and $x\subset y$ and for no larger $k'$ is there a $y$ satisfying these conditions. The function $L$ allows us to conveniently fill out a two-dimensional number grid. The grid is assumed to have as many places in the horizontal direction as there are terms in the clustering and $K$ places in the vertical direction. The grid is initially filled with zeros. The grid is filled out progressively from bottom to top. At level zero we take each cluster $x_i$ in the order they are numbered and working from left to right in the grid we place $i$ in the grid as many times as there are points in cluster $x_i$. This string of $i$'s then represents the cluster $x_i$ in the grid. For any cluster $x_j$ at a higher level we use the relation $L(x_j) = x_k$ to note that $x_k$ has already been placed on the grid as a string of $k$'s at a lower level. We then look above this string of $k$'s at $x_j$'s level and working left to right find the first grid point filled with 0. We then fill as many grid points at that level with $j$ as there are terms in cluster $x_j$. In this way every cluster is represented in the grid at its level with a string of numbers of length the size of the cluster. Further the clusters are organized vertically so that each cluster at a level above zero is placed over the larger cluster from which it was derived by splitting.

Based on the number grid just described it is now possible to create a graphical representation of the clustering results. We scan the grid from left to right looking down from the top level and record the first non-zero number we see moving from the top down for each horizontal position. We end this process when we reach a horizontal position where all the numbers at all levels are zero. As a result we will have a sequence $\{i_r\}$ of indices where $r$ represents the horizontal position and $i_r$ the number of a cluster that was placed on the grid. The sequence $\{i_r\}$ is composed of short runs of the same index representing a cluster. If the cluster represented was of size less than 20 we replace the numbers in the grid with the color blue at that location. If the cluster is of size 20 or larger, but not all the grid points for the cluster appear in the sequence $\{i_r\}$ we replace the numbers in the grid with green. This represents a part of a cluster of size greater than 100 that splits to produce a cluster or clusters at a higher level. If the cluster is of size 20 or larger and all the grid points for the cluster appear in the sequence $\{i_r\}$, we replace the numbers in the grid with the color red. This appears as a red bar in the graph. These red bars represent the most significant clusters and in order to improve their display and differentiation from each other we move each such red bar upward vertically as many levels as there are points in the cluster. The strongest clusters are represented by the blue and green peaks indicating many levels of splitting to obtain a cluster of size 100 or less.

## 3. Results and Evaluation

Evaluating the performance of topic modeling algorithms is a challenging task. It is challenging not only because manually created gold standards are required, but also because creating such gold standards is not a well-defined task. Results may vary depending on the goal of the task and be equally useful for their particular tasks. We evaluate our model based on its ability to compute meaningful topic terms.

### 3.1 Evaluating topic-term association with topic coherence measures

Topic Coherence measures score a topic by measuring the degree of semantic similarity between high scoring words in the topic. These measures capture the semantic interpretability of the topic based on topic subject terms. Recent studies have investigated several topic coherence measures in terms of their correlation with human ratings (Aletras and Stevenson 2013, Röder, Both et al. 2015). Two measures that have been demonstrated to correspond well to human coherence judgements are NPMI (normalized pointwise mutual information, also referred to as the UCI measure (Newman, Noh et al. 2010)), and the UMass measure (Mimno, Wallach et al. 2011).

Here we use the NPMI and the UMass coherence measures to evaluate the topic coherence on the *suicide* dataset. Our algorithm applied to the '*suicide*' dataset results in 302 topics. PDC computation is based on unigrams and bigrams. We evaluated our top scoring terms against those computed by LDA. The Mallet opensource tool (McCallum 2002) was used to run LDA on the *suicide* dataset using unigrams and bigrams and default parameters. Guided by the number of topics obtained by our method we ran LDA with the same number of topics as produced by PDC.

Table 1 presents the results based on UMass and NPMI coherence metrics respectively for the top 5, 10, and 20 topic words produced by PDC and LDA. Results demonstrate that top scoring terms computed by PDC achieve a better coherence score then those computed by LDA using the NPMI measure in all three settings. When using the UMass measure, LDA measures show better numbers than the PDC cluster terms.

Table 1: Comparative evaluation of PDC topics with LDA topics using the UMass (a) and NPMI (b) coherence metrics on the suicide dataset.

| UMass | Top 5 | Top10 | Top20 |
|---|---|---|---|
| PDC | -25.9908 | -201.642 | -1429.8 |
| LDA | **-19.1805** | **-104.29** | **-586.5** |

(a)

| NPMI | Top 5 | Top10 | Top20 |
|---|---|---|---|
| PDC | **7.98422** | **33.9968** | **135.381** |
| LDA | 6.49662 | 27.4303 | 106.706 |

(b)

To investigate this discrepancy, we examined the top 20 terms produced by PDC and LDA for each cluster. The number of unique tokens produced by PDC is 6,040. The number of unique tokens in LDA is 2,313. Moreover, we calculated the average document frequency of these terms. The average document frequency of the top 20 terms in the LDA clusters is 3,785, while the average document frequency of the top 20 terms in the PDC clusters is 143. We observed a very big difference in the document frequency of topic terms produced by these two methods, which demonstrate that the PDC algorithm identifies clusters of terms of a more specific nature than those identified by LDA topic terms. This analysis is illustrated in Figure 1.

This analysis may explain why the UMass measure numbers are higher for the LDA topic terms. These results highlight the differences between PDC and LDA. The normalization used in the NPMI corrects for the frequency difference between the two methods. Overall, the PDC algorithm has the advantage that we do not need to adjust the number of clusters. The optimal number is automatically found. Further the resulting topics, are more narrowly focused, which may be of value when researchers need a detailed view.

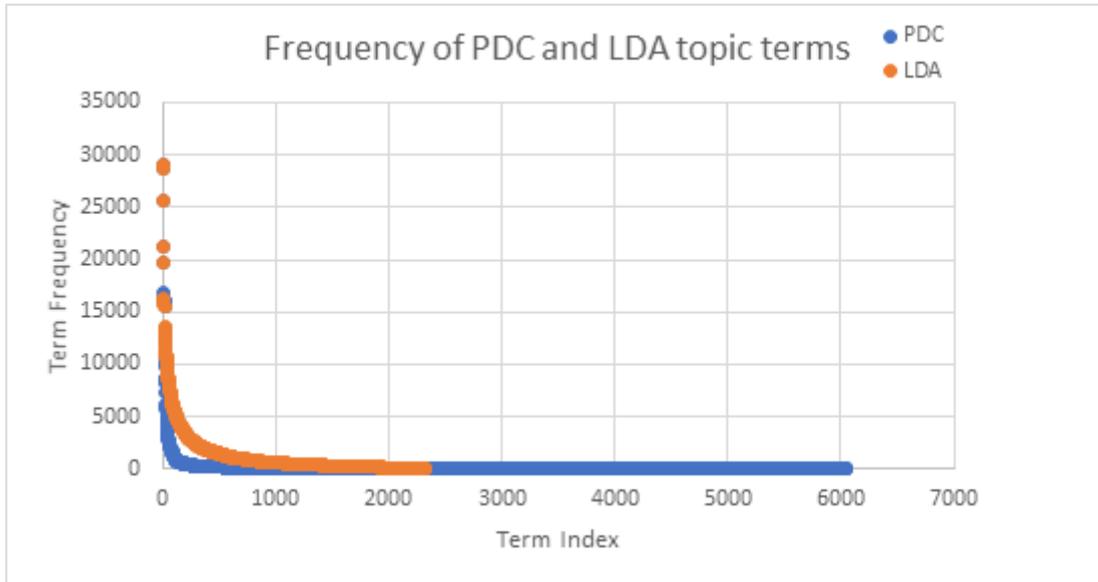

Figure 1. Document frequency of top 20 topic terms for the clusters identified by the PDC and the LDA algorithms, respectively.

### 3.2 The scope of Mental Health Illness and Suicide in PubMed Articles

In biomedical research, new knowledge is primarily presented and disseminated in the form of peer-reviewed journal articles. Searching through literature to keep up with the state of the art is a necessity for many individual biomedical researchers. In this work, we identify and study the set of PubMed articles related to *suicide* using the PDC clustering method.

When applied to the suicide literature, our topic analysis algorithm identified 302 topics, each topic being represented by topic terms along with the score. For each topic, we generated a topic name from either the top scoring MeSH term, or the top scoring bigram listed in the top twenty ranked topic terms. Then, all PubMed documents are scored with respect to each topic. Some of the largest topics are on "suicide risk factors", "mortality", "depressive disorders", "assisted suicide", "suicide prevention". The PDC algorithm also shows a clear partition of the literature where the research concentrates on "suicide gene", which is an important genetic therapy technique as a potential way of treating cancer and other proliferative diseases, as shown in Figure 2.

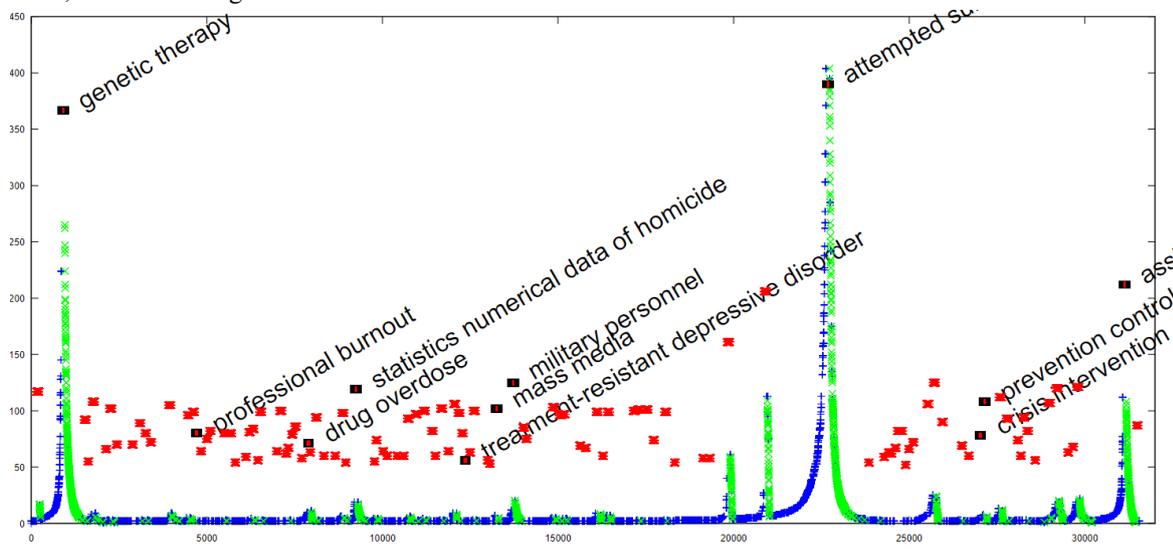

Figure 2 Graphical depiction of the output of the topic clustering algorithm. All peaks represent well-formed topic clusters. Clicking on each cluster reveals the set of terms that define that cluster, and the set of PMIDs that scored highest for that cluster. We have noted the cluster names depicted in Table 2.

Figure 2 is a graphical representation of all the term clusters identified using the PDC algorithm on the PubMed suicide literature. A closer inspection of these groups reveals coherent groups of terms, as we show in Table 2. In Table 2 we randomly selected ten clusters of terms from those depicted in Figure 2. As seen the number of terms varies. The most important terms in each cluster are shown in the table. For the full list of terms as well as the top scoring PubMed articles associated with them visit
https://www.ncbi.nlm.nih.gov/CBBresearch/Wilbur/IRET/MKXPOST/SUICIDE/suicide.svg.

Table 2 An illustration of top scoring term clusters from the distributional probabilistic clustering algorithm. For each cluster we show the number of terms, the cluster title and a list of top scoring terms.

| Topic Size (#of terms) | Topic Title | Top 10 terms |
|---|---|---|
| 100 | assisted suicide | assisted suicide // assisted // physician assisted // physician // legislation & jurisprudence of assisted suicide // euthanasia // right to die // personal autonomy // terminally ill // terminal care |
| 100 | mass media | media // mass media // imitative behavior // news // newspapers as topic // newspaper // media reporting // newspapers // celebrity // copycat // media coverage |
| 100 | attempted suicide | results // study // risk // suicidal // female // attempts // ideation // attempted suicide // conclusions // suicidal ideation |
| 100 | statistics & numerical data of homicide | homicide // firearms // statistics & numerical data of homicide // gun // mortality in gunshot wounds // statistics & numerical data of firearms // |
| | | legislation & jurisprudence of firearms // ownership // gun // firearm related |
| 100 | prevention & control of suicide | prevention & control of suicide // prevention // suicide prevention // preventing // preventing suicide // program evaluation // program // gatekeeper // health education // gatekeeper training |
| 100 | military personnel | military personnel // military // psychology of military personnel // army // soldiers // statistics & numerical data of military personnel // personnel // active duty // duty // combat |
| 73 | crisis intervention | crisis // crisis intervention // hotlines // telephone // callers // calls // methods of crisis intervention // telephone crisis // suicidal crisis // lifeline |
| 78 | professional burnout | professional burnout // burnout // psychology of professional burnout // epidemiology of professional burnout // prevention & control of professional burnout // workplace // complications of psychological stress // maslach burnout inventory // psychology of workplace // emotional exhaustion |
| 59 | drug overdose | drug overdose // overdose // opioid // mortality in drug overdose // opioid-related disorders // opioid analgesics // epidemiology of drug overdose // poisoning of opioid analgesics // psychology of drug overdose // prevention & control of drug overdose |
| 54 | treatment-resistant depressive disorder | ketamine // treatment-resistant depressive disorder // therapeutic use of ketamine // drug therapy of treatment-resistant depressive disorder // trd // depression trd // ketamine's psychology of treatment-resistant depressive disorder // acting // antidepressant effects // acting antidepressant |

To further help researchers make sense of this partitioning of the uses of the initial search term (suicide), clicking on a cluster from the graphical interface will bring up a pop-up window composed of two panels. On the left panel the whole list of cluster terms is shown. The ranking order corresponds to the score produced by the PDC algorithm. On the right panel we give the top scoring PubMed articles most associated with the cluster of terms on the left. Again, the ranking order reflects how well each article scores against the terms in the cluster on the left. This usage is depicted in Figure 3, where our selected cluster is the first one listed in Table 2.

## 4. Conclusion

In this study, we presented a probabilistic distributional clustering algorithm that can be used to describe a collection of terms pertaining to a major topic. Starting with a query term and selecting the set of documents returned from that query, this computational analysis allows the browsing of different topics that represent the usage of that term in the collection of documents. For example, applying the PDC algorithm on the suicide related literature in PubMed, we are able to see topics such as: attempted suicide, assisted suicide, suicide risks in youth and adolescents, suicide risks in military personnel but also suicide gene therapy. An important contribution of our work is the graphical literature analysis, which is a procedure that allows a global view of all topics and related documents as they are discovered in every stage of the PDC algorithm. Our presentation of the published literature as it partitions naturally following the probabilistic distributional approach, opens up new possibilities for researchers to examine the different aspects of a subject in the literature. The website is available at: https://www.ncbi.nlm.nih.gov/CBBresearch/Wilbur/IRET/MKXPOST/SUICIDE/suicide.svg.

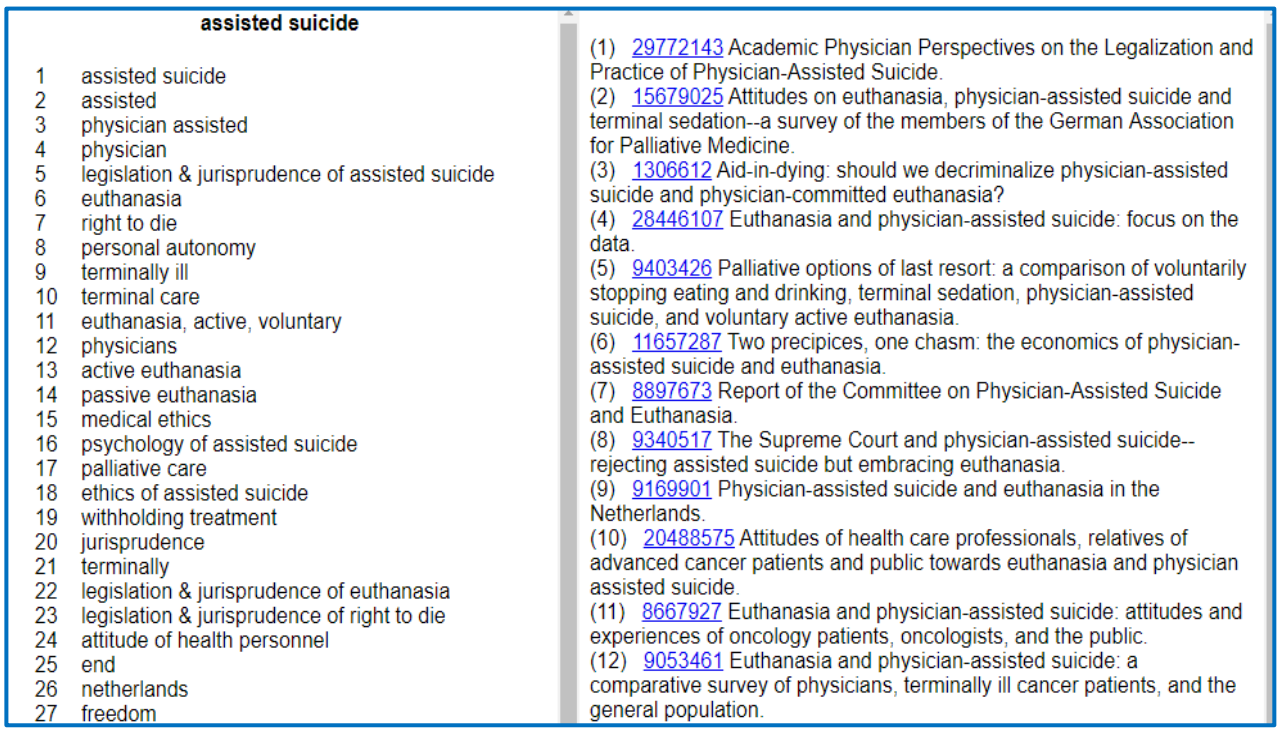

Figure 3 Clicking on a cluster in the graphical interface brings us a pop-up window showing the topic terms on the left, and the most related articles to that topic on the right. Clicking on each article links to the PubMed record. The figure demonstrates the topic "assisted suicide".

**Acknowledgements**

This research was supported by the NIH Intramural Research Program National Library of Medicine.

## Appendix 1. The PDC Algorithm in Pseudocode

Since we are dealing with a finite set $U$ it will be convenient to assume that the elements of $U$ are uniquely numbered with integer indices from 1 to $N$, i.e., $U = \{x_i\}_{i=1}^{N}$. Then if $\delta$ represents any partition function on $U$ we may define a second function $\bar{\delta}$ by setting $\bar{\delta}(x_i)$ equal to the smallest $j$ such that $\delta(x_i, x_j) = 0$. Then it is evident that $\bar{\delta}^{-1}(j)$ for any $j$, $1 \leq j \leq N$, is either empty or one of the clusters defined by $\delta$. Define

$$R = \{j \mid \bar{\delta}^{-1}(j) \neq \varnothing\}. \tag{19}$$

Then $\{\bar{\delta}^{-1}(j)\}_{j \in R}$ represents the clustering corresponding to the partition function $\delta$. While partition functions are convenient for some purposes, we do not find them convenient for computer implementation and we will instead work with integer arrays. Given a partition function $\delta$ we can define an integer array by

$$a[i] = j \Leftrightarrow x_i \in \bar{\delta}^{-1}(j). \tag{20}$$

Likewise, if $a$ is any integer array satisfying

$$0 \leq a[i] < N, \text{ for } 0 \leq i < N \tag{21}$$

we can define the corresponding partition function by

$$\delta(x_i, x_j) = \begin{cases} 0, & a[i] = a[j] \\ 1, & a[i] \neq a[j] \end{cases}. \tag{22}$$

Because equality is symmetric and transitive it is trivial to check that $\delta$ is a partition function. Henceforth we will restrict our attention to integer arrays satisfying (21).

To simplify our handling of the probability function $p$ it will be convenient to define

$$\sigma(i, j) = \log\left(\frac{p(x_i, x_j)}{1 - p(x_i, x_j)}\right) + factor, \ i \neq j \tag{23}$$

$$\sigma(i, j) = 0, \ i = j.$$

Here the *factor* represents the prior log odds ratio (17) and can be set to a value dictated by our goals. Consistent with (5) we seek an integer array $a$ that maximizes the expression

$$\sum_{i, j \ni a[i]=a[j]} \sigma(i, j). \tag{24}$$

Though the function $\sigma$ is symmetric, because the probability function is symmetric, it is convenient to store the whole $N \times N$ matrix of values on disk and access them by memory mapping. With these preliminaries we are ready to consider the optimization problem in the form (24).

In seeking to optimize $a$ it will be convenient to refer to $a[i]$ as the label of $i$. Then it is evident from (20) points have the same label if and only if they belong to the same cluster. In working with $a$ we will also require two additional arrays of the same length as $a$ which are defined on labels by

$$\begin{aligned} a\_cnt[j] &= \|\{i \mid a[i] = j\}\| \\ a\_sum[j] &= \sum_{a[s]=j=a[t]} \sigma(s,t). \end{aligned} \tag{25}$$

For any $j$, $0 \leq j < N$ which are not labels we require $a\_cnt[j] = a\_sum[j] = 0$. We will also need arrays $a'$, $a'\_cnt$, and $a'\_sum$ which are of the same length and meaning as the $a$ arrays and are used to hold preliminary results in the computation and are copied to $a$, $a\_cnt$, and $a\_sum$ only when they improve on the values already in these arrays. These six arrays are globally defined and are not passed as arguments in functions or routines. Now much of the time when the algorithm is working it is only working on a subset of the points $i$, $0 \leq i < N$. Such a subset will be represented by an array $b[i]$ defined on an interval $0 \leq i < nb(\leq N)$ and satisfying $0 \leq b[i] < N$ and $i \neq j \Rightarrow b[i] \neq b[j]$. We begin by defining the basic single point optimization where each point is moved in turn to a different cluster if that can improve the overall score. As long as this process succeeds it is repeated up to 30 times. The limit of 30 is used to avoid the rare case when slight improvement is continuing, for big improvements are generally seen early in this process.

```
function singleOpt( nb, b){
    cti=0
    flag=1
    while(flag){
        flag=0
        for(m=0; m<nb; m++){
            i=b[m]
            ic=a'[i]
            for(j=0; j<nb; j++)par[b[j]]=0
            for(j=0; j<m; j++){
```

```
                    k=b[j]
                    par[a'[k]]+=σ(i,k)
                }
                for(j=m+1; j<nb; j++){
                    k=b[j]
                    par[a'[k]]+=σ(i,k)
                }
                mx=par[ic]
                k=ic
                for(j=0; j<nb; j++){
                    if(mx<par[b[j]]){
                        mx=par[b[j]]
                        k=b[j]
                    }
                }
                if(k!=ic){
                    a'_sum[ic]-=par[ic]
                    a'_sum[k]+=mx
                    a'_cnt[ic]--
                    a'_cnt[k]++
                    a'[i]=k
                    flag=1
                }
            }
            if((++cti)>=30)break
        }
        sxm=0
        for(j=0;j<nb;j++)sxm+=a'_sum[b[i]]
        return(sxm)
    }
```

When the singleOpt function returns an improved sum we record this result.

```
recordMax(nb,b){
    for(i=0;i<nb;i++){
        k=b[i]
        a[k]=a'[k]
        a_cnt[k]=a'_cnt[k]
        a_sum[k]=a'_sum[k]
    }
}
```

The singleOpt function and recordMax are used together to optimize over a subset of points.

```
function localOpt(nb,b,xsum){
    randomly shuffle the order of elements in the array b
    ysum=singleOpt(nb,b)
```

```
        if(ysum>xsum){
            recordMax(nb,b)
        }
        return(ysum)
}
```

The function localOpt is used by the basic splitting algorithm which attempts to improve the score of the elements associated with a particular cluster by trying different splittings of the cluster. In the process of trying different splittings it is helpful to have a routine that sets the values in the $a'\_sum$ array.

```
setSum(nb,b){
    for(i=0;i<nb;i++)a'_sum[b[i]]=0
    for(i=0;i<nb;i++){
        for(j=i+1;j<nb;j++){
            if(a'[b[i]]==a'[b[j]]){
                a'_sum[a'[i]]+=σ(b[i],b[j])
            }
        }
    }
}

basicSplit(nb,b){
    if(nb==1){
        k=b[0]
        a[k]=-k-1  (labels are converted to a negative integer for those clusters that cannot be split)
        a_cnt[k]=1
        a_sum[k]=0
        return
    }
    //Find the most negative points in the set
    bs=0
    for(i=0;i<nb;i++)neg_sum[i]=0
    for(i=0;i<nb;i++){
        for(j=i+1;j<nb;j++){
            bs+=xx=σ(b[i],b[j])
            if(xx<0){
                neg_sum[i]+=xx
                neg_sum[j]+=xx
            }
        }
        ord[i]=i
    }
    xs=bs
    Sort the array ord so that neg_sum[ord[i]] is in increasing order
    for(i=0;i<10;i++){
        if(neg_sum[ord[i]]<0){  // Process starting with most negative points
            cxt++
```

```
//Order points reverse to how postive they are to neg point
sm=0
for(j=0;j<nb;j++){
    sm+=pos[j]= σ(b[i],b[j])
    ord2[j]=j
}
Sort the array ord2 so pos[ord2[i]] is in descending order
//Count how many positive points there are
tp=0
while(pos[ord2[tp]]>0)tp++
//sm tells how the neg point relates to all others
if(sm<0){  //create a split and test it
    if(i==0)n=1
    else n=0
    for(j=0;j<nb;j++){
        a'[b[j]]=b[n]
        a'_cnt[b[j]]=0
        a'_sum[b[j]]=0
    }
    a'_sum[b[n]]=xs-sm
    a'_cnt[b[n]]=nb-1
    a'_cnt[b[i]]=1
    a'[b[i]]=b[i]
    bs=localOpt(nb,b,bs)
}
if(tp>0){  //create splits and test them
    k=1;
    while(k<=tp){
        for(j=0;j<nb;j++){
            a'[b[j]]=b[ord2[k]]
            a'_cnt[b[j]]=0
        }
        a'[b[i]]=b[i]
        a'_cnt[b[i]]=k+1
        for(j=0;j<k;j++)a[b[ord2[j]]]=b[i]
        a'_cnt[n]=nb-k-1
        setSum(nb,b)
        bs=localOpt(nb,b,bs)
        if(k*2<=tp)k=k*2
        else if(k<tp)k=tp
        else k=tp+1
    }
}
        }
    }
}
//relabel so each label is in its cluster
flag=0
```

```
    for(i=0;i<nb;i++){
        if(a_cnt[b[i]]){
            lbl[flag]=b[i] //collect labels (count > 0)
            cnt[flag]=a_cnt[b[i]] //collect counts
            sum[flag++]=a_sum[b[i]] //collect sum for each class
        }
    }
    for(i=0;i<nb;i++)ord[a[b[i]]]=b[i] //map labels to one of the points with that label
    if(flag==1){ //If cannot divide then label switched to neg.
        j=-ord[lbl[0]]-1
        for(i=0;i<nb;i++){
            k=b[i]
            a[k]=j
            a_cnt[k]=0
            a_sum[k]=0
        }
    }
    else {
        for(i=0;i<nb;i++){
            k=b[i]
            a[k]=ord[a[k]]
            a_cnt[k]=0
            a_sum[k]=0
        }
    }
    for(i=0;i<flag;i++){
        a_cnt[ord[lbl[i]]]=cnt[i]
        a_sum[ord[lbl[i]]]=sum[i]
    }
}
```

The basicSplit routine is called repeatedly on the clusters of size greater than *thr* until no such cluster can be further split. This is accomplished by the masterSplit function.

```
function masterSplit(thr){
    flag=1
    while(flag){
        flag=0
        m=0;
        while(m<N){
            while((m<N)&&((a[m]<0)||(a_cnt[m]<=thr)))m++
            if(m<N){
                flag=1
                n=a[m]
                nb=0
                for(i=m;i<N;i++){
                    if(a[i]==n)b[nb++]=i
```

```
                }
                basicSplit(nb,b);
            }
        }
    }
    flag=0
    for(m=0;m<N;m++){//convert all labels back to nonnegative numbers
        if(a[m]<0){
            a[m]=-a[m]-1
            flag++
        }
    }
    return(flag)
}
```

The masterSplit function can be called once the arrays *a*, *a_cnt*, and *a_sum* are initialized. This is simply done by creating a single cluster using the oneCluster routine.

```
oneCluster(){
    for(i=0;i<N;i++){
        a[i]=0
        a_cnt[i]=0
        a_sum[i]=0
        b[i]=i
    }
    a_cnt[0]=N
    for(i=0;i<N;i++){
        for(j=i+1;j<N;j++){
            a_sum[0]+=σ(i,j)
        }
    }
}
```

The masterSplit function will split any cluster of size greater than *thr* if it can. However, there is no guarantee that a cluster can be split regardless of its size. The ability to split depends on the value of *factor* in (23). By sufficiently decreasing *factor* we can always split a cluster. Our general approach to clustering is to begin with *factor* = 0 and incrementally decrease *factor* by a value *del* until all clusters remaining are of size less than or equal to *thr*. This is implemented in the superSplit routine.

```
superSplit(thr,del){
    factor=0
    oneCluster()
    masterSplit(1)
    printFile()
    while(masterSplit(thr)>0){
        printFile()
         factor-=del
```

```
        }
}
```

Here the printFile routine is included to indicate one may want to obtain output regarding the clustering at each level used in the processing. The definition of printFile we leave up to the user as it is application dependent.